\documentclass[letterpaper, 10 pt, conference]{IEEEtran}
\IEEEoverridecommandlockouts

\usepackage{cite}
\usepackage{amsmath,amssymb,amsfonts}
\usepackage{algorithmic}
\usepackage{graphicx}
\usepackage{textcomp}
\usepackage{xcolor}
\usepackage{wrapfig}
\usepackage{balance}

\usepackage{todonotes}

\def\BibTeX{{\rm B\kern-.05em{\sc i\kern-.025em b}\kern-.08em
    T\kern-.1667em\lower.7ex\hbox{E}\kern-.125emX}}
\begin{document}

\title{How does a robot's social credibility relate to its perceived trustworthiness?}

\author{\IEEEauthorblockN{Patrick Holthaus}
\IEEEauthorblockA{\textit{School of Physics, Engineering, and Computer Science} \\
\textit{University of Hertfordshire}\\
Hatfield, United Kingdom\\
\texttt{p.holthaus@herts.ac.uk}}
}

\maketitle

\section{Introduction}

This position paper aims to highlight and discuss the role of a robot's \textit{social credibility} in interaction with humans. In particular, I want to explore a potential relation between social credibility and a robot's acceptability and ultimately its trustworthiness. I thereby also review and expand the notion of social credibility as a measure of how well the robot obeys social norms during interaction~\cite{Menon2019} with the concept of conscious acknowledgement.

\section{Social robot acceptability}

Humans interaction comprises social signals that nonverbal exchange theories often argue to enrich the communication channel with additional information~\cite{Patterson1982}. Likewise, robots are often designed in a sociable way and programmed to exhibit social competence with the aim to improve the quality and effectiveness of an interaction~\cite{Esposito2016}.
A wide range of experiments in the research field of human-robot interaction (HRI) evaluate particular elements or combinations of robot appearances and designs or verbal and nonverbal behaviours with regard to the human's social perception of the robot.

Many different factors affecting people's social perception of a robot have already been identified, such as the naturalness of a robot's movements~\cite{Lichtenthaler2012}, expressiveness and vulnerability of a robot~\cite{Martelaro2016}, emotions~\cite{Saerbeck2010}, and proximity~\cite{Koay2007}.
Social behaviours can provide a robot with the ability to establish and maintain social relationships by using natural cues, and expressing and perceiving emotions~\cite{fong2003survey}. To be acceptable by humans, such social behaviour must be meaningful and congruent~\cite{Hegel2011}, appropriate to the social role that it is expected to fulfil~\cite{koay2014social}, and continuously provides appropriate signals for the entire duration of an interaction~\cite{Holthaus2021}.

Typically, the acceptability of a robot's social functions is verified using questionnaire scales like  \textit{Godspeed}~\cite{bartneck2008measuring} or the \textit{Robot Social Attribute Scale} (RoSAS)~\cite{RoSAS} that \textit{implicitly} measure the participants' perception of the robot in different experimental conditions. That is, participants normally rate general robot attributes like reliability, competency, happiness, or scariness.
Qualitative evaluation, open questions, and objective measures like participant compliance with robot requests help to further reason about user ratings and preference.
As a result, appropriate robot sociability, i.e. design and behaviours that lead to the robot's acceptability, are usually acknowledged subconsciously by human participants and exact reasons are often interpreted using additional data.

\section{Trust in social robots}

Trust is one of the fundamental factors for a successful cooperation between humans and robots \cite{Hancock2011,Salem2015Trust,Rossi2017b}. The research field of HRI employs and develops several definitions of trust that originate in psychology and are then adapted to interaction with robots. Most of these definitions (e.g.~\cite{Lewis1985}) include cognitive and affective factors that contribute people's to assessment of an robots reliability by its actions and behaviours. On this basis, people can form an emotional connection with robots on the assumption that both (human and robot) are positively invested in the success of the interaction and relationship \cite{McAllister1995,deVisser2020}.
Consequently, the social acceptability of robot behaviours can heavily influence a robot's perceived trustworthiness.

\section{Social robot credibility}

\begin{figure}
  \centering
  \includegraphics[width=\linewidth]{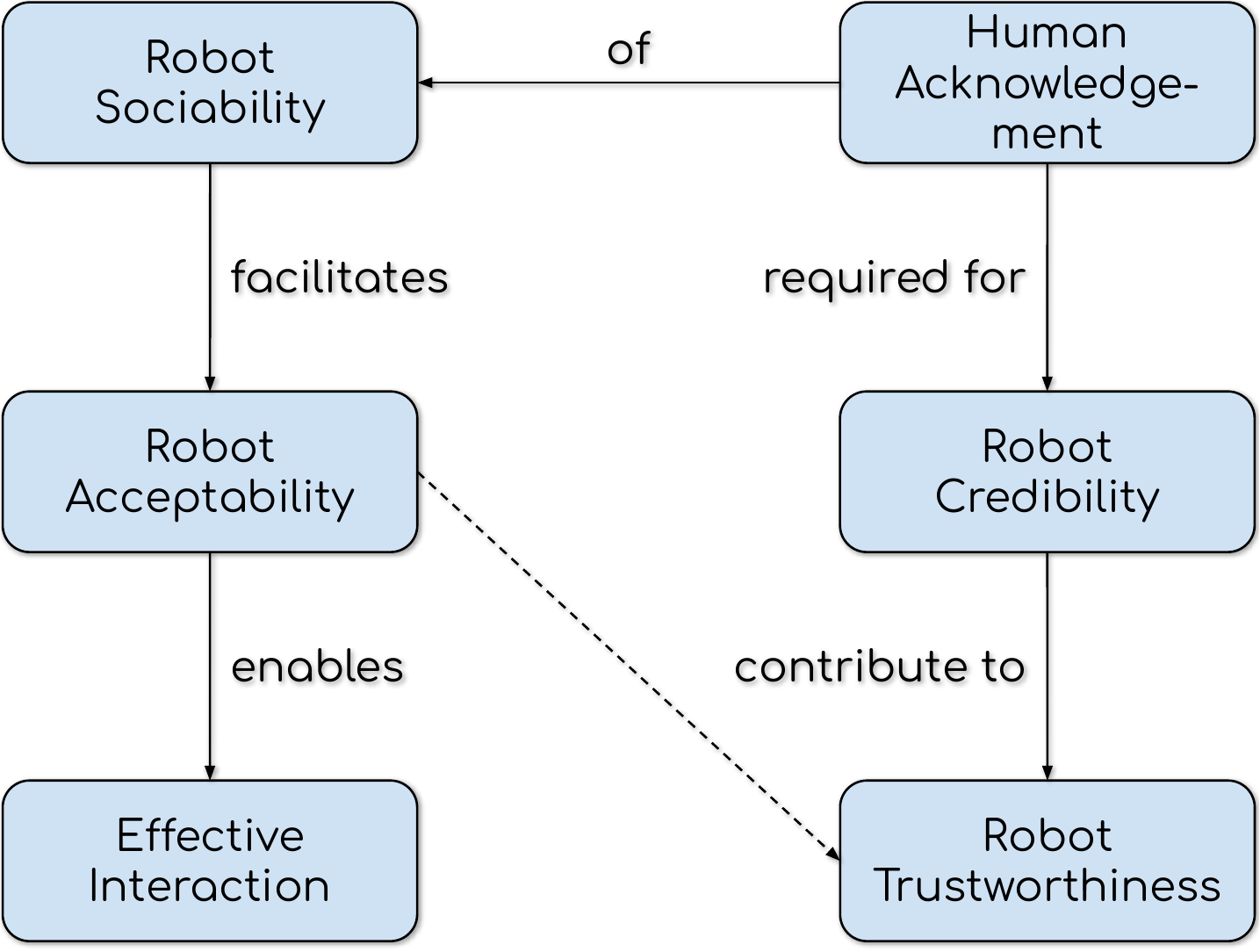}
  \caption{Robot sociability: How a human's conscious and unconscious acknowledgement might lead to acceptability, credibility and trustworthiness}
  \label{fig:authors}
\end{figure}

A robot's \textit{social credibility} has been introduced as a measure of "how well it obeys the social norms relevant to its environment"~\cite{Menon2019}. As such, the term is closely related to the notion of socially acceptable behaviour but not entirely identical. Behaviours can be socially credible but not acceptable, e.g. when a robot is impolite or interrupting a conversation. At the same time, they can also be acceptable but not credible, e.g. when a robot is pretending to empathise with human emotions.
The concept of social credibility aims to quantify how believable and authentic a robot's social behaviours are. It thereby explicitly considers the context of existing social norms where robot behaviour can be credible by staying within the range of behaviours that would be expected from a human interaction partner.

The definition further implies that a robot's social credibility is always evaluated with regard to the environment and robot itself.
That is, the credibility of a behaviour is tied to a situation with one specific robot in a particular environment as expectations on normative behaviour might be in a different in other situations.
Much like behaviours that are socially not acceptable, socially incredible robots are more likely to be perceived negatively by people during an interaction.
For example, a robot might be perceived as having less authority \cite{Holthaus2019} when acting as a safety monitor.

\section{Influence of credibility on trustworthiness}
In addition to the above, it is noteworthy that if a certain robot behaviour is supposed to be credible, a human needs to acknowledge it as deliberately exhibited by the robot with the specific purpose to facilitate social interaction.
Social credibility as the humans conscious acknowledgement of a robot's sociability might be an important factor that contributes to a robot's trustworthiness.
If we are able to capture whether users recognise or not recognise a robot's social actions as such we could further develop the conscious part of a humans mental model about a robot that affect its perceived trustworthiness.
At the same time, the acceptability of robot behaviours contributes to the person's mental model of the robot, albeit to the subconscious part, cf. Fig.~\ref{fig:authors}.

The credibility of a robot's social behaviour might also have an important role when the human's trust in the robot changes or is lost. Robot actions that highlight the positive investment in the interaction by demonstrating its engagement in social behaviour that do not serve any obvious other function have the potential to be used as repair mechanisms. Explicit social signals like a communicative gaze, blinking, facial expression, gesture, or a colloquial utterance might have a positive effect in such situations even if they are interrupting the otherwise acceptable interaction.

It is not yet entirely clear how to quantify a robot's social credibility.
I argue that we need to capture the human's conscious acknowledgement of the robot's sociability to successfully measure social credibility and draw better conclusion about a robot's perceived trustworthiness.
It might therefore be worthwhile to develop and adopt subjective and objective measurement standards that allow conclusion about a robot's social credibility as a complementary method to sociability in HRI experiments.

\balance  

\bibliographystyle{IEEEtran}
\bibliography{SCRITA2021-social-trust}

\end{document}